\documentclass{article}

\usepackage{PRIMEarxiv}
\usepackage{changes}

\usepackage[utf8]{inputenc} % allow utf-8 input
\usepackage[T1]{fontenc}    % use 8-bit T1 fonts
\usepackage{hyperref}       % hyperlinks
\usepackage{url}            % simple URL typesetting
\usepackage{booktabs}       % professional-quality tables
\usepackage{amsfonts}       % blackboard math symbols
\usepackage{nicefrac}       % compact symbols for 1/2, etc.
\usepackage{microtype}      % microtypography
\usepackage{lipsum}
\usepackage{fancyhdr}       % header
\usepackage{graphicx}       % graphics
\graphicspath{{media/}}     % organize your images and other figures under media/ folder

\usepackage{algorithm}
\usepackage{algorithmic}
\usepackage{amsmath}
\newcommand{\alg}{NPS-SQA }
\newcommand{\algnospace}{NPS-SQA}
  
\title{What If: Generating Code to Answer Simulation Questions}

\author{
  Gal Peretz \\
  Technion, \\Israel Institute of Technology \\
  \texttt{sgalprz@cs.technion.ac.il} \\
   \And
  Kira Radinsky \\
  Technion, \\Israel Institute of Technology \\
  \texttt{kirar@cs.technion.ac.il} \\
}

\begin{document}
\maketitle

\begin{abstract}
Many texts, especially in chemistry and biology, describe complex processes. We focus on texts that describe a chemical reaction process and questions that ask about the process's outcome under different environmental conditions. 
    To answer questions about such processes, one needs to understand the interactions between the different entities involved in the process and to simulate their state transitions during the process execution under different conditions. A state transition is defined as the memory modification the program does to the variables during the execution.
    We hypothesize that generating code and executing it to simulate the process will allow answering such questions. We, therefore, define a domain-specific language (DSL) to represent processes. We contribute to the community a unique dataset curated by chemists and annotated by computer scientists. The dataset is composed of process texts, simulation questions, and their corresponding computer codes represented by the DSL.
   We propose a neural program synthesis approach based on reinforcement learning with a novel state-transition semantic reward.
   The novel reward is based on the run-time semantic similarity between the predicted code and the reference code. This allows simulating complex process transitions and thus answering simulation questions.
   Our approach yields a significant boost in accuracy for simulation questions: 88\% accuracy as opposed to 83\% accuracy of the state-of-the-art neural program synthesis approaches and 54\% accuracy of state-of-the-art end-to-end text-based approaches.
\end{abstract}

% keywords can be removed

\section{Introduction}
Many texts, especially in chemistry and biology, describe complex processes. These texts describe a chemical reaction process and require answering questions asking about the process's outcome under different environment conditions. 
To answer these questions, one needs to understand the interactions between the different entities involved in the process and to simulate their state-transitions during the process execution under other conditions. 
Consider the following example:
\begin{quote}
\small
    \textbf{A mixture of 50 g (0.42 mole)} of p-aminobenzonitrile, 146 ml of concentrated hydrochloric acid, and 106 ml of H2O \textbf{was heated at 80 for 20 minutes} and then cooled to 0°. ... The hexane was cooled and the resulting precipitate was filtered and air dried to \textbf{yield 16 g (18\%) of p-(5-acetyl-2-furyl) benzonitrile}. ... Overheat the mixture to \textbf{83 degrees} will result in \textbf{9\% decrease in the outcome of p-(5-acetyl-2-furyl) benzonitrile}."

    \textbf{Question}: What will be the output of the process, if after 10 minutes of heating the mixture we increase the heat by 0.5 degree per minute?
\end{quote}
This example describes chemical reactions between several components that yield 16 grams of  p-(5-acetyl-2-furyl) benzoni-trile. A model that attempts to answer the question following the text needs to extract the right information about the temperature and the heating duration described in the text, and to understand the output of the iterative process described in the question. Finally, it needs to perform a numerical computation and to check if the temperature reached the threshold of 83 degrees or not. We refer to these type of questions as \emph{Simulation Questions}, as they require to simulate a process under other environment conditions. 

In recent years, deep learning models have achieved state-of-the-art performance for question-answering
(QA) and reading comprehension tasks. Pre-trained models like Roberta \cite{Roberta} and XLNet \cite{YangDYCSL19} are end-to-end models that achieved high performance on QA datasets like SQuAD \cite{RajpurkarZLL16} and SWAG \cite{SWAG}.
Although these models achieved impressive results on various QA tasks, they mostly focus on span-based datasets, where the answer to the question must be a span from the text itself, or questions that require simple commonsense reasoning.
For simulation questions one needs the ability to simulate the process at hand in addition to performing a numerical computation that would yield the final process product.

Semantic parsing approaches, whose goal is to convert a natural language text to a logical form, showed superior results for inference of simple processes-based texts datasets (\cite{BerantSCLHHCM14,KiddonZC16,YaoLGSS19,TandonDSCB19}) that require simple numeric reasoning (\cite{Dua2019DROP, amini-etal-2019-mathqa}). 
However, simulation questions require understanding the complex process' dynamics in additional to numerical and scientific commonsense reasoning capabilities. Consider our aforementioned example. The question requires to simulate an iterative process of heating until reaching a certain threshold. This requires to mimic a loop and if-clause.
In this work, we both present a unique dataset (SimQA) for complex processes and an algorithmic framework to answer simulation questions. Our approach attempts to predict the question's answer by modeling the process described in the text as a computer code simulating the process's progress.

\begin{figure}
    \centering
    \includegraphics[width=0.6\linewidth]{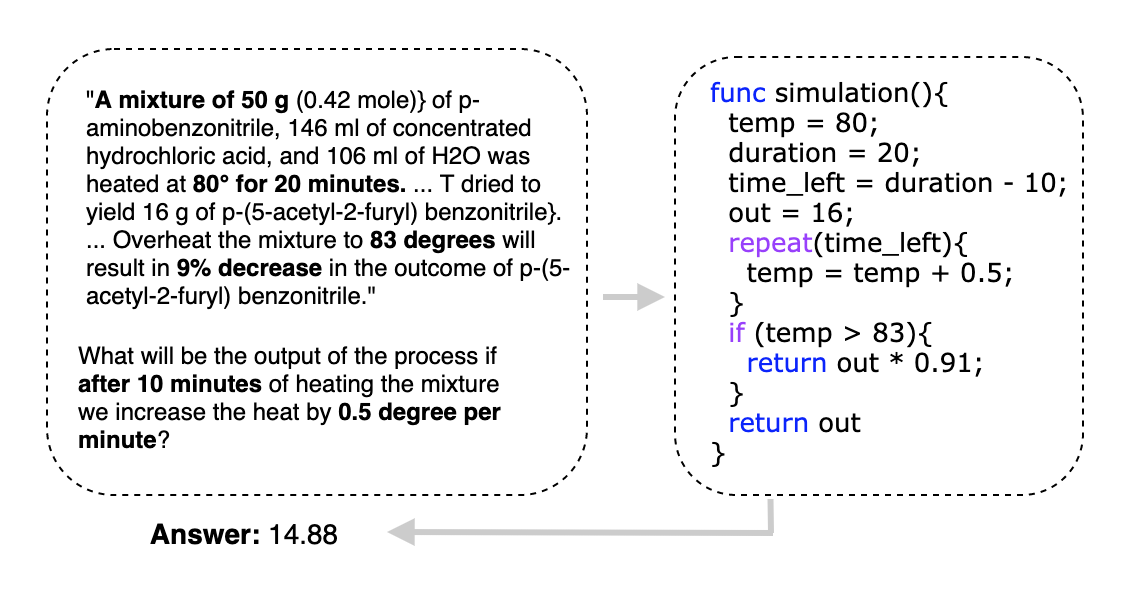}
    \caption{\small{A text of a process and the corresponding code.}}
    \label{fig:code_example}
\end{figure}

We create SimQA by extending the US patents dataset (\cite{Lowe2017}) -- a well-known dataset in the domain of chemistry. Chemists were assigned to create questions based on texts that describe chemical experiment, and computer scientists were assigned to annotate the questions and the texts with a corresponding program that if executed, answers the question correctly (Figure~\ref{fig:code_example}).
For the program annotation, we define a domain-specific language (DSL). It allows to represent a code that simulates a process described in a text given the different environment settings described in the question.
All of the answers in the SimQA dataset are numerical values that can be inferred from the text and the question by understanding the process's components interactions and the computation that will lead to the correct answer.
  \begin{figure}[h!]
    \includegraphics[scale=0.6]{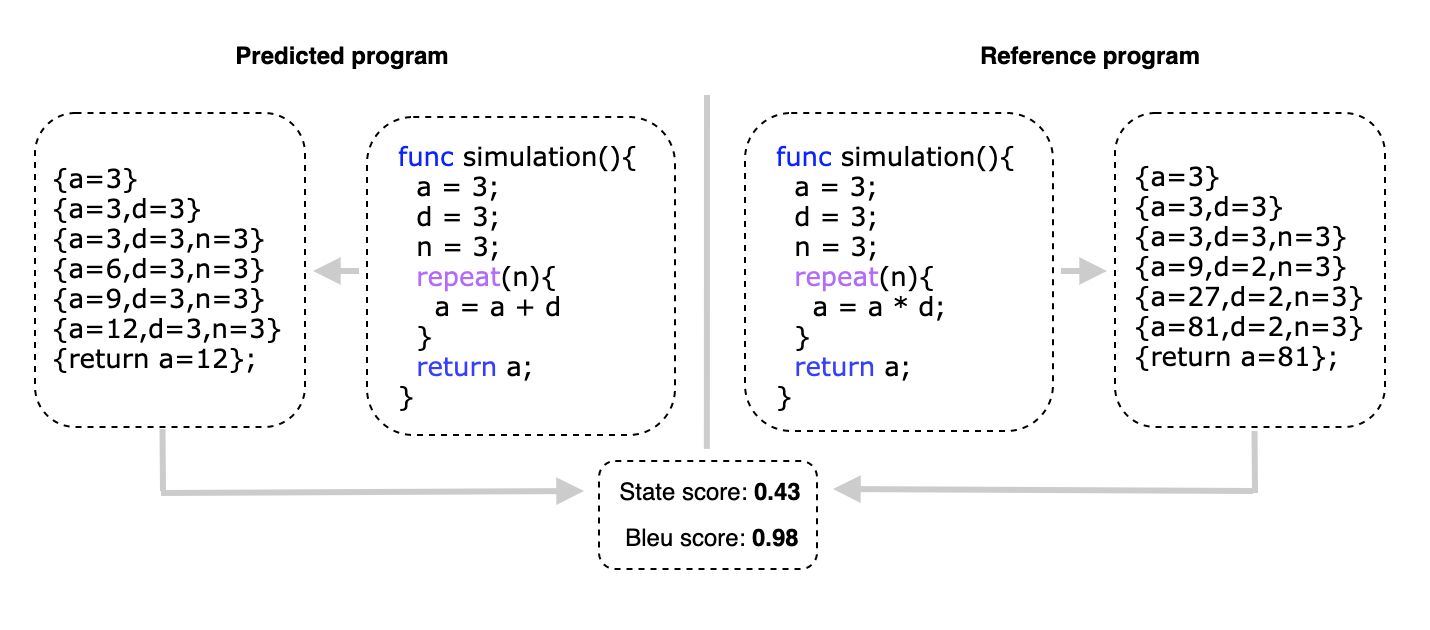}
    \caption{\small{State and BLEU score for program synthesis.}}
    \label{figure:BLEUvsState}
  \end{figure}
  \subsection{Syntactic Neural Program Synthesis}
  \label{sec:mle}
  \label{sec:mle}
We propose an algorithm we call Neural Program Synthesis for Simulation QA (\algnospace) that given a text and a question, attempts to learn to produce the corresponding code represented by the DSL. We treat the problem as a translation problem and learn to translate from text to code. 
The complexity of the generated DSL that contains loops, if-clauses, etc., requires a more nuanced optimization than simple end-to-end translation models. 
Unlike previous approaches that focused on optimizing only the syntax of the target program, our approach leverages the run-time information to estimate semantic similarity between the predicted and the reference programs. We devise a reinforcement learning approach and during optimization create programs.  Those are then run and compared to the oracle code run-time variable state. We refer to the latter as semantic code similarity.
We study the trade-off between the semantic and syntactic similarity rewards, and provide insights to the type of questions that \alg outperforms state-of-the-art QA models.

The contribution of our work is fourfold:
(1) We present a novel dataset for simulation questions of complex processes. It contains reference programs that can be executed to infer the right answers to the questions. As far as we know, this is the first question-answering dataset that requires deep understanding of the scientific processes' dynamics and numerical commonsense reasoning capabilities. We contribute it to the community for further research \footnote{\protect\url{https://drive.google.com/file/d/1jzfAWjjmgGZtUVVke9TB9zEn4tM_7ThM/view?usp=sharing}}.
(2) We define a domain-specific language (DSL) that is more concise than full featured programming languages like Python, but expressive enough with the ability to imitate the state of the process by defining variables, simulate iterative process using loops and examine multiple possibilities using if-conditions.
(3) We propose a novel reward function based on the state transition of the predicted and the reference programs to model the dynamics of the process. We use policy gradient training to train our model by rewarding it based on both the syntactic similarity and run-time similarity of the execution processes. We publish our code for the benefit of the community \footnote{\protect\url{https://github.com/galprz/SimQA}}. 
    We perform an empirical analysis comparing \alg and state-of-the-art deep-learning QA approaches showing significant boosts of 30\%+ in accuracy compared to pretrained end-to-end question-answering models and 5\%+ for neural program synthesis approaches trained on SimQA.

\section{Related Work}
\label{chap:related}
% Question answering is a broad field that can be broken down into subdomains. Models that excel in one subdomain may perform poorly in others. 
Recently, pretrained models like ELMo(\cite{ELMO}), BERT(\cite{devlin-etal-2019-bert}), XLNET(\cite{YangDYCSL19}) and Roberta(\cite{Roberta}) have shown state-of-the-art performance on various QA tasks like SQUAD (\cite{RajpurkarZLL16}), RACE (\cite{RACE}), GLUE (\cite{WangSMHLB19}) etc. However, these models perform poorly on tasks that require logical or numerical commonsense~\cite{lin-etal-2020-birds}. On the other hand semantic parsing approaches transform the text to a logical form that facilitates inference. These methods showed superior performance on tasks that require understanding of processes' dynamics ~\cite{berant-etal-2013-semantic,dasigi-etal-2019-iterative,LiangBLFL17,MEMAug,JiaL16, DongL16, LapataD18} and numerical commonsense reasoning~\cite{amini-etal-2019-mathqa, chenLYZSL20}. Semantic parsing approaches have two variations, weakly supervised approaches \cite{berant-etal-2013-semantic,dasigi-etal-2019-iterative,MEMAug,LiangBLFL17,MinCHZ19, chenLYZSL20} and fully supervised approaches. Our model is a fully supervised approach.

% Weakly supervised approaches attempt to overcome the shortage of an oracle program by using maximizing marginal likelihood \cite{berant-etal-2013-semantic,dasigi-etal-2019-iterative}, reinforcement-learning combined with memory augmentation method \cite{LiangBLFL17,MEMAug} or iterative expectation maximization \cite{LiangBLFL17, MinCHZ19, chenLYZSL20} for bootstrapping. 
Fully supervised approaches use sequence-to-sequence architectures in a fully supervised manner to optimize the maximum marginal likelihood of logical language that the computer can interpret given as a label (``the oracle''). For example, if the logical language is a computer program code, then the oracle is the predicted code or the target abstract syntax tree \cite{JiaL16, DongL16, LapataD18} corresponding to the text.
Recent example of a semantic parsing model that achieved state-of-the results on well known benchmarks like Drop~\cite{Dua2019DROP} and MathQA~\cite{amini-etal-2019-mathqa} is Neural Symbolic Reader(NeRd) \cite{chenLYZSL20} that has an encoder-decoder architecture and can be used as a weakly or fully supervised model.

The aforementioned approaches deal with limited target DSL without variables, if branches and loops. 
Consider a representative question from MathQA:
``A train running at the speed of 48 km / hr crosses a pole in 9 seconds . What is the length of the train ?''. The logical form is ``multiply(divide(multiply(48, const 1000), const 3600), 9)''.  Unlike the above example, chemistry texts contain a description of iterative processes and conditional events, therefore, often require more expressive DSL. Consider the complex process example in Fig.~\ref{fig:code_example} that requires the notion of loops and if-clauses in addition to numerical comprehension of multiple entities.

The complexity of the generated DSL requires a more nuanced optimization. Unlike previous approaches that focused on optimizing the syntax of the target program, 
our approach leverages the \emph{run-time information} of the oracle program during optimization. 
We devise a reinforcement learning approach and during optimization compare the predicted code to the oracle code run-time variable state. Our reward is therefore composed of both syntactic and run-time similarity. 
This helps the model to converge even when the DSL is complex and the search space is large.
% Nevertheless, most of the successes of semantic parsing approaches deal with limited target DSL without variables, if branches and loops. Often, chemistry texts contain a description of iterative processes and conditional events, therefore, require more expressive DSL. Our approach is fully supervised and leverage run-time information of the oracle program. This helps the model to converge even when the DSL is complex and the search space is large.

\section{Problem Definition}
\label{chap:problem}
Let $\Sigma$ denote an input text, $\Gamma$ denote a simulation question, and $y \in R$ to denote the correct answer. The question-answering task is to find $\pi:(\Sigma,\Gamma)\rightarrow y$, which maps the text and the question to $y$, the correct answer.
In this work, we focus on simulation questions.
A simulation question is a question that asks someone to imagine what might happen. 
Formally, we focus on problems that can be solved by the auxiliary task of $\pi':(\Sigma,\Gamma) \rightarrow \lambda$,  which maps the text and the question to a code, s.t. $\lambda()=y$, i.e., when executing $\lambda$ it produces $y$.

\section{Methodology}
\label{chap:prelims}
\subsection{Domain Specific Language}
We start by defining a language $\lambda$, s.t. a text can be matched to it.
We aim to define a concise version of a full-featured programming language to reduce the size of the search space of possible programs thus potentially leading to better convergence.
We define this domain-specific language (DSL) to resemble c-syntax language, but with a less complex syntax while still retaining the capability of defining variables, ``if'' branches, and loops to manipulate a variables' state.
The following is the grammar definition of the new language:

\begin{footnotesize}
\begin{align*}
  &Prog\ p := \text{func simulation()}\ \{\ s\ \}\\
  &Stmt\ s := \text{repeat(}n\text{)}\ \{\ s\ \}\ \ |\ \ \text{if}(c)\{\ s\ \}\\
  &\indent \indent \indent \indent|\ \text{return}\ e\ ;\  |\ \ a; \ \ |\ \ s_1;s_2\  \\
  & Cond\ c\ :=\ identifier\ op\_c\ e\\ 
  &Assign\ a\ :=\ identifier\ =\ e\ \\  
  &\indent \indent \indent \indent |\ identifier = e\ op\ e\\
  &e\ :=\ identifier\ |\ n\ | \ \text{-}n\ |\ r\ |\ \text{-}r\\
  &op\_c :=\ <\ |\ >\ |\ >=\ |\ <=\ |\ !=\ |\ ==\\
  &op\ :=\ \ \ +\ \ |\ -\ |\ *\ |\ \ \backslash\ \ \ |\ \backslash \backslash\\
  &identifier\ :=\ [a\text{-}zA\text{-}Z][a\text{-}zA\text{-}Z0\text{-}9\_]*\\
  &n\ :=\ [0\text{-}9]\text{+}\\
  &r\ := \ [0\text{-}9]\text{+.}[0\text{-}9]\text{+}
\end{align*}
\end{footnotesize}
We use this grammar to derive an abstract syntax tree (AST) to validate the syntax of the programs.
We provide tools to translate the generated code to python code to allow code execution.  
\subsection{Neural Program Synthesis for Question-Answering}
Assume we have a labeled dataset of $(\Sigma_i,\Gamma_i,\lambda_i)$. One can refer to the problem as a translation task, i.e., given a text $\Sigma_i$ and a question $\Gamma_i$ generate the corresponding code, $\lambda_i$, that answers the question $\Gamma_i$.
The common approach \cite{WangLS17, rajpurkar-etal-2018-know} is to use an encoder-decoder architecture that maximizes the likelihood (MLE) of the corresponding text. In our setting that corresponding text is the reference program.
Recently, it has been shown that an additional step of optimization, applied after the model is trained, can significantly boost results \cite{RLSeqOptimization,RLBLEUOptimization2, Seq2SeqWithRL}. 
Usually, reinforcement learning is applied during this optimization step.
We adopted the architecture of the encoder-decoder model and used Transformers (as suggested in Attention is all you need paper \cite{AttentionIsAllYouNeed}) and train the model using a cross-entropy loss.
We treat the trained decoder from the previous step (optimized for MLE) as a model-free agent and further optimize the code generation process using reinforcement-learning. We use beam search to sample $N$ possible programs that have the highest sequence probability, calculated by the multiplication of the probability for each token that constructs the program. We denote the reward function as $Q:(\lambda,\hat{\lambda}) \rightarrow R$ and apply it on the predicted programs. We take the $S$ programs with the highest value, as determined by $Q$. Because the target function is non-smooth we cannot use it as a loss function and backpropagate the gradient thus, 
we use the REINFORCE \cite{Seq2SeqWithRL,OnPolicyGradients} algorithm to optimize the reward using the following loss function:
 \begin{align}
 \mathcal{L}
  =-Q( \lambda_{i},\hat{\lambda_i}) log(p(\hat{\lambda_i}| \Sigma_i, \Gamma_i, \lambda))
\end{align}

We gather a batch size of $B$ generated code sequences. For each sequence the loss function is calculated. We average the loss across the $B$ sequences. The result is back-propagated to update the weights of the encoder-decoder model. 
Algorithm \ref{alg:reinforce} illustrates the training:
After training the encoder-decoder model using the cross-entropy loss (line 2) we iterate over all training examples. For each example a beam-search is applied to return the $N$ sequences of the highest probabilities (line 4). The algorithm then selects the top sequences based on the reward $Q$ (lines 5-7). The final loss is calculated for the batch (lines 8-11). 

\begin{algorithm}
\small
    \caption{Training}
        \hspace*{\algorithmicindent} \textbf{Input: $N,S,B$}
    \begin{algorithmic}[1]
      \STATE $batch \gets$ []
      \STATE $model\ \gets$ pre-train MLE model
      \FOR{$(\Sigma_i,\Gamma_i,\lambda_i)$ in dataset}
        \STATE $n\_top \gets$ beam\_search($model,\Sigma_i,\Gamma_i, N$)
        \STATE $s\_top \gets $[$(Q(\lambda,\hat{\lambda}),\hat{\lambda})$ for $\hat{\lambda}$ in n\_top])
        \STATE $s\_top \gets$ sort(s\_top)  \# sort by $Q$ value
        \STATE $batch \gets$ batch + s\_top[:S]  \# take top s
        \IF{length($batch$) $=$ $B$}
          \STATE $\mathcal{L}
          \gets -\frac{1}{B}\sum_{i \in batch}{Q_i log(p(\hat{\lambda_i}| \Sigma_i, \Gamma_i))}$
          \STATE $\mathcal{L}.backward()$
          \STATE $batch \gets $[]
        \ENDIF
      \ENDFOR
    \end{algorithmic}
    \label{alg:reinforce}
  \end{algorithm}
  We consider a reward function, $Q$, that will measure both the the syntactic similarity between $\lambda_i$ and $\hat{\lambda_i}$, and the semantic similarity. We denote the syntactic similarity as $Q_{syntactic}$ (Section~\ref{sec:mle}) and the semantic as $Q_{semantic}$ (Section ~\ref{sec:semantic}).
  We then set the final $Q$ as the weighted sum of the syntactic and semantic scores:
  \begin{align}
  \label{eq:q}
  Q=\gamma Q_{syntactic}+(1-\gamma) Q_{semantic}
  \end{align}

  Recent research has shown that an additional step of BLEU-score optimization, applied after a model is trained, can produce dramatic improvements in performance \cite{RLSeqOptimization,RLBLEUOptimization2, Seq2SeqWithRL}. 
  Intuitively, training the encoder-decoder architecture using cross-entropy loss for each token optimizes the max likelihood estimation (MLE) for each token separately, while BLEU-score optimization optimizes the sequence as a whole. The BLEU-score can give a quantitative estimation of the syntactic similarity of the predicted code with the reference code. Therefore, we define: $Q_{syntactic}= BLEU(\hat{\lambda_i},\lambda_i)$.

  \subsection{Semantic Neural Program Synthesis}
  \label{sec:semantic}
  
  Unlike translation-based approaches, we hypothesize that when generating code, it might not be sufficient to only require syntactic similarity to the reference program.
  A small syntax mistake can lead to a wrong output, syntax errors, or run-time errors. Consider the example presented in Figure~\ref{figure:BLEUvsState}. The left program simulates an iterative process that eventually outputs the sum of an arithmetic series while the right program calculates the sum of a geometric series. The two programs are close syntactically but semantically different. Therefore, we aim to capture the semantics of the code execution alongside the syntax. 
  
  A computer program stores data in variables. The contents of these variables during the code, at any given point in the program's execution, is called a \emph{program's state}. We define semantic similarity of code programs as the similarity of the program's states during code execution.
  Formally, we define $\hat{S_t}=\{\hat{v_1}^t,\hat{v_2}^t,...,\hat{v_n}^t\}$ as the state of the predicted program and $S_t=\{v_1^t,v_2^t,...,v_n^t\}$ as the state of the reference program at time $t$. Note that if the program didn't define $v_i$ yet we treat it's value as $\epsilon$, a special symbol that define that the variable has not been used.
  
  We define $\hat{I}=(\hat{\sigma_1},\hat{\sigma_2},...,\hat{\sigma_m})$ and $I=(\sigma_1,\sigma_2,...,\sigma_k)$ as the instructions of the predicted program and the oracle program correspondingly. An instruction is a line of code that can change the state of the memory.  We use function $E:(S_t,\sigma_t)\rightarrow S_{t+1}$ to execute the instruction in time t and compute the next state at time t+1. We use these definitions to define semantic reward function that rewards each state that is aligned with the reference code state.
  
  Unlike $Q_{syntactic}$ that measures the syntactic similarity between the generated code and $\lambda_i$, $Q_{semantic}$ should capture the similarity of the states of the the generated code and the reference program $\lambda_i$.
  We aim to integrate the run-time semantic estimation into the training process.
  To evaluate the similarity of the code states at each step, $\lambda_i$ and the generated code are executed and a reward is calculated:
  
  \begin{footnotesize}
  \begin{align*}
    &T_{max} = max(|I|,|\hat{I}|),T_{min} = min(|I|,|\hat{I}|)\\
    &Q_{semantic}=\frac{1}{T_{max}} \sum_{t=0}^{T_{min}}
    \begin{cases}
      1, &\forall i \in [n],\\
       &v_i^t\in E(S_t,\sigma_t),\\
       &\hat{v_i}^t\in E(\hat{S_t},\hat{\sigma_t}),\\
       &v_i^t=\hat{v_i}^t \\
      0,              & \text{otherwise}
    \end{cases}
  \end{align*}
  \end{footnotesize}
  After the execution of both programs, the reward is estimated by the number of matched state transitions.
  Algorithm \ref{alg:Algorithm1} presents the semantic reward algorithm.
  The input to the algorithm is the reference code and the predicted code. The state transitions after each line are recorded during code execution. A check is then performed if the two programs modified the variables in the same way and returned the same values.
  \begin{algorithm}
  \small
    \caption{Semantic Reward}
    \hspace*{\algorithmicindent} \textbf{Input: $\lambda_i$, $\hat{\lambda_i}$}\\
    \begin{algorithmic}[1]
      \STATE $\textit{r\_state} \gets \textit{execute\_and\_get\_state($\lambda_i$)}$\\
      \STATE $\textit{p\_state} \gets \textit{execute\_and\_get\_state($\hat{\lambda_i}$)}$\\
      \STATE $correct\_states \gets 0$
      \STATE $T_{min} \gets min(len(r\_state),len(p\_state))$
      \FOR{$t=0$ to $T_{min}$}
          \IF{all(r\_state[t] = p\_state[t])}
            \STATE $correct\ states \gets correct\_states + 1$
          \ENDIF
      \ENDFOR
      \STATE $T_{max} \gets max(len(r\_state),len(p\_state))$
      \RETURN $correct\_states/T_{max}$
    \end{algorithmic}
      \label{alg:Algorithm1}
  \end{algorithm}

  Note that we require that all the variables will have the same values for each execution timestamp, but we can think on other variations that allow equality only for part of the variables, or allow the state to be equal on different stages of the execution and not only in the same timestamp. We leave these experiments to future research.

  \section{Experimental Methodology}
  \label{chap:experimental-methodology}
  \subsection{Datasets}
  The chemical reactions' dataset contains descriptions of reactions between chemicals from the US patents published between 1976-2016.
  Each reaction is annotated with the corresponding components, their quantities and actions that affected the process outcome.
  Consider this example from the U.S. Patent dataset:
  \begin{quote}
      "A mixture of \colorbox{green}{50 g} (0.42 mole) of \colorbox{yellow}{p-aminobenzonitrile}, \colorbox{green}{146 ml} of \\ \colorbox{yellow}{concentrated hydrochloric acid}, and \colorbox{green}{106 ml} of \colorbox{yellow}{H2O} was \colorbox{red}{heated} at \textbf{80c} \\ for \textbf{20 minutes} and then \colorbox{red}{cooled} to \textbf{0°}. ... \colorbox{yellow}{The hexane} was \colorbox{red}{cooled} and the resulting precipitate was \colorbox{red}{filtered} and air \colorbox{red}{dried} to yield \colorbox{green}{16 g} (18\%) of \colorbox{yellow}{p-(5-acetyl-2-furyl) benzonitrile}."
      \label{labledText}
  \end{quote}

The example describes the process's components (yellow), their quantities (green) and the actions (red) that were taken to cause the components to interact with each other to yield 16g of p-(5-acety-2-furyl) benzonitrile. We used the US patents dataset to create the SimQA datasets -- question-answering datasets on texts that contain descriptions of chemical processes and reactions. 
  Chemists were asked to curate a set of question and answers over entities, actions and quantities.
  We provide two datasets based on those questions, SimQA-v1 and SimQA-v2.
  For SimQA-V2, the chemists were only allowed to create questions about the reaction described. An example of such question on the above text:
    ``How many milliliters of water we need to add to yield 25 grams of the product if we want to preserve the same ratio between them?''.
    
For SimQA-V1, the chemists were allowed to add additional plausible information about the reaction to allow creating more advanced questions. The context is added to the dataset.
An example of such context: ``If the temperature exceed 100 degrees when heating the mixture it causes the water to be evaporated at a rate of 2.3 milliliters per minute.'', and the corresponding question:
``How many milliliters of water should we add if we increase the temperature by 2 degrees per minutes for 10 minutes and we want to cancel the effect of the evaporation?''
  
Besides the true answer, we generate additional three possible answers for each question to match the multiple-choice task requirements. This allows us to compare against state-of-the-art models for the end-to-end approach. We generate the three wrong answers randomly, but close in value to the true answer see example in appendix A. Note that our model doesn't use these answers and instead try to predict the answer directly from the question and the reference code. The datasets contain 6187 training examples (3158 in SimQA-V1 and 3029 in SimQA-v2) and 1089 test examples (555 in SimQA-V1 and 534 in SimQA-V2).
  
\section{Baselines and Metrics}
To evaluate the value of solving the auxiliary problem of finding the program that answers the question instead of answering the question directly, we compare against state-of-the-art question-answering models: Roberta~\cite{Roberta}, XLnet~\cite{YangDYCSL19} and Longformer~\cite{Longformer}.We use Roberta and XLnet as baselines because recently they reached state-of-the-art performance on various question answering datasets, and we use Longformer because our task often requires to learn long dependencies between the chemical components. 
We use pre-trained models and fine-tune them on the SimQA training datasets.
Because these models perform well on QA tasks when the answer is in the text itself or with a multi choice setup, Beside the right answer we generate another three wrong answer that are close numerical to the right one. we give these baselines the advantage of choosing the right answer from the 4 possibilities while our model should output the right answer directly without seeing the 4 choices.

To compare our approach to other semantic parsing approaches we used Seq2Prog\\~\cite{amini-etal-2019-mathqa}, an encoder decoder architecture based on LSTM models which leverages attention mechanism to decode the right program, and NeRd \cite{chenLYZSL20} (its fully-supervised version) that also uses encoder(the "reader") decoder(the "programmer") architecture based on BERT and LSTM models that achieved state-of-the-are performance on MathQA \cite{amini-etal-2019-mathqa} and Drop~\cite{Dua2019DROP} datasets -- both are a well-known benchmark for commonsense inference and logical reasoning tasks. Note that these baselines have the same task as our model to - to predict the answer to the question given the context, question and reference code. 
For the \algnospace, after hyper-parameters tuning (separate validation set), we chose to use a beam width of 32 sequences. We set $N=8$, $S=4$ and $B=32$.
All the models were evaluated using the accuracy metric, i.e., the percentage of correct answers.

\section{Experiments and Results}
\label{chap:experimental-methodology}
In this section, we present the main empirical results of \alg compare to the other baselines. We run our experiments using Tesla V100 GPU with 16GB of memory.
\section{Main Result}
Table \ref{tab:main} presents the main results of our approach on the two SimQA datasets. Statistically significant results are shown in bold. We observe that \alg outperforms current state-of-the-art methods significantly with a boost of 33\% - 36\% in accuracy against the end-to-end models. This shows the merit of using code generation as an auxiliary task. Additionally, 
we observe a boost of 4\% - 5\% compared to other semantic parsing approaches. This shows the merit of code execution as part of the run-time similarity between the oracle program and the predicted program during the training process.
The SimQA-v1 dataset contains questions that require an understanding of the process and chain of events that lead to the final product. As we show in \ref{strength} this is the strength of \alg which explain the better performance of \alg on SimQA-v1 over SimQA-v2.
%TODO: Add explanations why better performance for v1 vs v2
  
  \begin{table}[ht]
  \centering
  \caption{Main results}
  \small
  \begin{tabular}{| c| c| c |}
  \hline
    \textbf{Model} & \textbf{SimQA v1} & \textbf{SimQA v2} \\
  \hline
   Roberta & 0.5233 & 0.5112 \\  
  \hline
   XLnet & 0.5421 & 0.5102 \\
   \hline
   Longformer & 0.5291 & 0.4991 \\
   \hline
    Seq2Seq & 0.8013 & 0.7921 \\
   \hline
   Seq2Prog & 0.8279 & 0.8021 \\
   \hline
   NeRd & 0.8261 & 0.8041 \\
   \hline
   \alg  & \textbf{0.8801} & \textbf{0.8431} \\
   \hline
  \end{tabular}
  \label{tab:main}
  \end{table}
  \newpage
  \section{The Syntactic and Semantic Trade-off}

\begin{figure}[h!]
        \centering

        \includegraphics[scale=0.5]{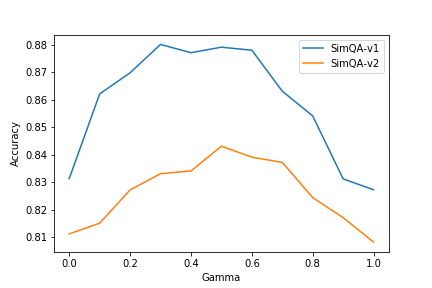}
      \caption{\small{The performance of the model with the syntactic-semantic reward for different $\gamma$ values.}}
      \label{figure:gamma}
  \end{figure}

    The reward function of $Q$ is composed of both syntactic and semantic similarity (Eq. \ref{eq:q}).
    To better understand the right balance between the syntactic reward, $Q_{syntactic}$, and the semantic reward, $Q_{semantic}$, we experimented with different values of $\gamma$.
    Figure \ref{figure:gamma} indicates that the optimal value for $\gamma$ is around $0.5$ on both datasets. This strengthens the hypothesis that both syntactic and semantic similarity help improve the performance.
    
    \subsection{Syntactic Reward}
    We observe that for $\gamma=1$, where only the syntactic similarity is used during the reward, the model reaches $0.8253$ and $0.8082$ for SimQA-v1 and SimQA-v2 respectively which match the performance of other approaches based on syntactic similarity.
    
    %TODO: Add explanation why better performance than for regular synatic methods
    \subsection{Semantic Reward}
    We observe that for $\gamma=0$, where only the semantic run-time code similarity is used during the reward, the model reaches $0.8313$ and $0.8112$ for SimQA-v1 and SimQA-v2 respectively.
    This shows that even if we use only state-score optimization after MLE optimization we can boost the performance of the model.
  \subsection{Syntactic Neural Program Synthesis}
  \label{sec:mle}
  \label{sec:mle}
    \subsection{Strengths of \algnospace}
    \label{strength}
    Mapping the text and the question to code as an intermediate step helps the model to find the chain of events that lead to the correct output. We observed that this approach performs well when dealing with questions that require some complex numerical calculation or execution of iterative loops and if branches. Formally, we define a \emph{Complex Simulation Question} as a question that requires a code that includes at least one if branch and one loop branch. This code needs to modify the state of the variables more than 15 times.
     An example of such a question is:
    \emergencystretch 1.5em
    \begin{quote}
    \small
        \textbf{Context}:
        A mixture of 5-hydroxy-1-tetralone (2.00 g) and sodium hydroxide (493 mg) was warmed in ethanol (40 mL) to 50° C. ....  If the temperature passes 56 degrees when heating \textbf{ethanol} for \textbf{more than 5 seconds} it will result in a loss of  \textbf{2.04 grams} in the final products \textbf{for each additional second}.
    
        \emergencystretch 1.5em
        \textbf{Question}: How many grams the process would yield if we increase the temperature when heating ethanol by 2 degrees every 2 seconds for 16 seconds?
    \end{quote}
    We extracted a subset of 443 examples with these characteristics and compared the performance of \alg to Roberta, NeRd and Seq2Prog models after training on the other examples.
    
    Table \ref{tab:complex} shows that our approach outperforms these models by a large margin on those questions that are prevalent when dealing with texts that describe scientific processes.
    
    \begin{table}[ht]
    \centering
    \caption{\small{Performance for complex questions}}
    \small
    \begin{tabular}{| c| c| c |}
    \hline
      \textbf{Model} & \textbf{Accuracy - Complex QA} \\
    \hline
     Roberta & 0.4271 \\  
     \hline
      Seq2Prog & 0.8183  \\
     \hline
      NeRd & 0.8153  \\
    \hline
      \alg & \textbf{0.8703}  \\
     \hline
    \end{tabular}
    \label{tab:complex}
    
    \end{table}
    
    \subsection{Limitations of \algnospace}
    The code generation approach works well on complex questions that involve state transitions or complex numeric calculations. However, if the question is simple the performance boost compared to baselines is lower. We define a \emph{Simple Simulation Question} as a question that requires a code that includes no more than one if branch and no loop branches. This code is allowed to change the state no more than 5 times.
    For example: 
    \begin{quote}
    \small
        \textbf{Context}: 
        ...prepared as described in example 7A, are refluxed at \textbf{160° C.} for 16 hours in a mixture of 9 ml of dimethylformamide, 1.3 ml of 85\% formic acid and 0.5 ml concentrated hydrochloric acid. After the reaction m ... with benzene. \textbf{4 g of the title product are obtained}.  If the temperature \textbf{exceed 175 degrees when heating dimethylformamide it hurts the process}

        \emergencystretch 1.5em
        \textbf{Question}: What will be the outcome of the process in grams if the temperature \textbf{increases by 14 degrees}?
    \end{quote}
    We gathered 361 simple questions examples and compared the performance of \alg against the same baselines after training on the other examples.
    For such questions, \alg still outperforms the performance of the state-of-the-art end-to-end model but matches the performance of the semantic parsing approaches (see Table \ref{tab:simpleqa}) \begin{table}[ht]
    \centering
    \caption{\small{Performance for simple questions}}
    \small
    \begin{tabular}{| c| c| c |}
    \hline
      \textbf{Model} & \textbf{Accuracy - Simple QA} \\
    \hline
     Roberta & 0.6027 \\  
      \hline
     Seq2Prog & 0.8215 \\  
     \hline
     NeRd & 0.8247 \\  
    \hline
      \alg & 0.8231  \\
     \hline
    \end{tabular}
    \label{tab:simpleqa}
    
    \end{table}

\section{Conclusions}
In this work, we explored the problem of simulation question-answering. 
We presented an approach that tackles this problem by generating a code that simulates the process described in the text and the environment conditions of the question. 
We automatically construct a dataset of descriptions of chemical processes. Chemists then curated possible questions and answers for those process descriptions. Each description and question were mapped to a codes of the process described.
We train an encoder-decoder model that optimizes both for syntactic similarity and semantic code similarities. 
Once the code is generated it is executed and outputs the answer to the question.
We show that such approach significantly boosts performance by more than 30\% in accuracy as compared to the state-of-the-art QA.
We study the semantic-syntactic trade-off and identify that both are needed for successful question-answering and process simulation. Additionally, we identify that questions that require complex simulation of process, with if-branches and loops, especially benefit from the intermediate task of code generation and execution. We also show the potential of explainability of the model using the generated code.
In future work, we wish to study the integration of external facts and world knowledge to code generation.
In this work, all the information that is needed to answer the question can be found in the text itself without the need to combine knowledge from external sources. For example, the model does not know that the boiling temperature of water is 100 C if it was not written in the text explicitly. Additional such knowledge can significantly improve broad simulation-question answering.    
%Bibliography
\bibliographystyle{unsrt}  
\bibliography{references}

\end{document}